\documentclass[conference]{IEEEtran}
\IEEEoverridecommandlockouts
\usepackage{cite}
\usepackage{amsmath,amssymb,amsfonts}
\usepackage{algorithmic}
\usepackage{graphicx}
\usepackage{textcomp}
\usepackage{xcolor}
\begin{document}

\title{Generalize Then Personalize: Wearable-Based Heart Rate Forecasting Under Scarce Ruck-March Data
}

\author{
\IEEEauthorblockN{
Muntaqim Ahmed Raju\IEEEauthorrefmark{1},
Ruizhe Ma\IEEEauthorrefmark{1},
John W. Ramsay\IEEEauthorrefmark{2}, and
David J. Cornell\IEEEauthorrefmark{1}
}
\IEEEauthorblockA{\IEEEauthorrefmark{1}University of Massachusetts Lowell, Lowell, MA, USA}
\IEEEauthorblockA{\IEEEauthorrefmark{2}U.S. Army DEVCOM Soldier Center, Natick, MA, USA}
\IEEEauthorblockA{
Email: \{MuntaqimAhmed\_Raju, Ruizhe\_Ma, David\_Cornell\}@uml.edu\\
John.W.Ramsay4.CIV@army.mil
}
}

\maketitle

\begin{abstract}
Heart rate (HR) forecasting may provide advance information about near-future cardiovascular response, potentially supporting physiological monitoring and workload-management decisions during ruck marching. However, robust model development is constrained by the difficulty of collecting large-scale ruck-march datasets, creating a practical scarce-data problem in which models must generalize across participants while adapting to individuals from limited personal data. To address this problem, we benchmark population-only training (POT), personalized-from-scratch training (PFST), and population-pretrained fine-tuning (PPFT)  for HR forecasting during loaded ruck marching using data from eight active-duty U.S. Army medics across thirteen sessions. LSTM, Attention-LSTM, and CNN-LSTM backbones were evaluated against a persistence baseline at 30~s and 60~s horizons. Across all neural architectures and horizons, PPFT yielded lower mean MAE, RMSE, and endpoint MAE than both POT and PFST. Participant-level paired analysis after averaging across three training seeds showed that, relative to PFST, PPFT reduced trajectory MAE by 2.22--4.30~bpm and endpoint MAE by 1.24--2.88~bpm. All six trajectory-MAE comparisons remained significant after Holm adjustment ($p_{\mathrm{adj}}=0.0469$), while four of six endpoint-MAE comparisons remained significant after adjustment. At 30~s, persistence achieved the lowest MAE (\textbf{$3.555 \pm 1.274$~bpm}), while CNN-LSTM PPFT achieved the lowest RMSE (\textbf{$4.669 \pm 1.916$~bpm}); LSTM PPFT achieved the lowest endpoint MAE (\textbf{$5.093 \pm 2.365$~bpm}). At 60~s, Attention-LSTM PPFT achieved the lowest MAE (\textbf{$5.573 \pm 1.939$~bpm}) and RMSE (\textbf{$7.134 \pm 2.260$~bpm}), while CNN-LSTM PPFT achieved the lowest endpoint MAE (\textbf{$7.553 \pm 2.357$~bpm}). These results suggest that population pretraining followed by participant-specific fine-tuning may provide a practical strategy for scarce-data wearable HR forecasting.
\end{abstract}


\section{Introduction}
\label{sec:introduction}

Ruck marching is a core military task involving prolonged load carriage and sustained cardiovascular demand \cite{spiezia2025loaded, earl2020effects}. These demands can contribute to fatigue, musculoskeletal strain, and overuse injury risk \cite{bloch2024effect, knapik2004soldier, almonroeder2021effects}, motivating continuous wearable monitoring during high-load activity \cite{williamson2015estimating, bustos2021non, friedl2018military}. While conventional monitoring describes the current physiological state, heart rate (HR) forecasting extends this capability by estimating near-future cardiovascular response, providing anticipatory information that may support physiological monitoring and workload-management applications \cite{koltun2023physiological, de2025real, nazaret2023modeling}.

Wearable-based HR modeling has been studied in broader exercise and free-living contexts, including running, fitness tracking, and large-scale consumer wearable datasets. Prior work has shown that HR dynamics can be modeled using activity signals, workout history, user embeddings, and individualized cardiovascular signatures \cite{nazaret2023modeling,hallgrimsson2018learning,ni2019modeling,spathis2021self,ludwig2018measurement}. A recurring pattern across these studies is that population-scale learning and personalization are complementary: population data can support transferable physiological representations, while participant-specific information can account for differences in fitness, recovery state, resting HR, and cardiovascular response. Generalization and personalization can therefore be viewed as sequential components of a generalize-then-personalize strategy.

Loaded ruck marching, however, differs from the exercise settings most commonly studied in wearable HR forecasting. It combines external load carriage, prolonged marching, task-specific movement patterns, and accumulated fatigue \cite{earl2020effects, bloch2024effect, feigel2025wearable}. Cardiovascular and perceptual strain during load carriage can also vary with carried load, heat, repeated bouts, recovery state, and workload accumulation \cite{vine2024metabolic,shuhada2020effects,arcidiacono2023peak,klughardt2025performance}. These factors can shift the HR--workload relationship relative to ordinary running or free-living walking, making direct transfer from general exercise models uncertain. At the same time, collecting ruck-march data requires military-relevant protocols, load carriage, trained participants, prolonged field sessions, wearable instrumentation, and substantial logistical effort \cite{bloch2024effect, faghy2022physiological, feigel2025wearable,koltun2023physiological}. Consequently, scarcity in this setting is primarily a participant- and session-level constraint: many overlapping forecasting windows can be generated from a small dataset, but participant diversity, repeated sessions, and truly independent participant-specific adaptation data remain limited.

This creates the central model-design question of this study: when
participant-specific ruck-march data are scarce, should HR forecasting
rely on population-only training (POT), personalized-from-scratch
training (PFST), or population-pretrained fine-tuning (PPFT)? POT
leverages cross-participant data but does not explicitly adapt to the
target participant's physiological response. PFST provides
participant-specific modeling but must learn both general HR dynamics
and participant-specific behavior from limited personal data. PPFT
provides a hybrid alternative by first learning shared HR--workload
structure from other participants and then adapting that representation
using data from the target participant.

Although population learning and personalization are well motivated by
adjacent wearable HR literature, to the best of our knowledge, POT,
PFST, and PPFT have not been directly benchmarked for ruck-march HR
forecasting under scarce-data conditions. Accordingly, this study
evaluates a generalize-then-personalize strategy for wearable HR
forecasting during loaded ruck marching. We compare POT, PFST, and PPFT
at 30~s and 60~s forecast horizons using leave-one-participant-out
evaluation. LSTM, Attention-LSTM, and CNN-LSTM models are evaluated as
distinct temporal backbones to determine whether the effect of training
regime persists across model families. We further examine whether
trend--residual decomposition provides additional benefit to PPFT.

\noindent\textbf{The main contributions of this work are as follows:}
\begin{itemize}
    \item We formulate wearable HR forecasting during loaded ruck marching as a scarce-data personalization problem and, to the best of our knowledge, provide the first controlled benchmark of POT, PFST, and PPFT in this setting.
    
    \item We evaluate POT, PFST, and PPFT under participant-held-out, chronologically ordered testing at 30~s and 60~s horizons across three neural forecasting backbones and a persistence baseline, allowing the effect of training regime to be examined across architectures.
    
    \item We quantify the performance of PPFT relative to PFST, POT, and persistence using participant-level paired analysis across three training seeds, and assess trend--residual decomposition as a secondary PPFT ablation.
\end{itemize}

\section{Methodology}

\subsection{Dataset}

The dataset comprised wearable measurements from 8 active-duty U.S. Army medics (all male) across 13 ruck-march sessions, with each participant contributing one or two sessions. Each session followed a 6.4-km (4-mile) load-carriage protocol at approximately 6.4~km/h (4.0~mph) while carrying a 15.9-kg (35-lb) pack. Mean \(\pm\) SD participant characteristics were \(20.3 \pm 2.0\) years for age, \(176.5 \pm 6.0\) cm for height, and \(78.9 \pm 11.7\) kg for body mass. HR was measured using Polar H9 monitors, while Catapult Vector S7 devices provided global positioning system (GPS)- and accelerometer-derived movement and workload signals \cite{merrigan2023validation,ocak2025validity}. The synchronized input variables were HR ($\mathrm{bpm}$), velocity ($\mathrm{m/s}$), acceleration ($g$), and Player Load (arbitrary units; AU). Player Load is a cumulative wearable-derived measure of physical work computed from triaxial accelerations and is distinct from the fixed 35-lb external pack load.

\subsection{Data Preprocessing and Sequence Generation}

Raw wearable signals were recorded at 10~Hz and resampled to 1~Hz to reduce high-frequency variability and align all channels to a common sampling interval. Within each second, HR, velocity, and acceleration were averaged, whereas the final Player Load value was retained because Player Load is cumulative. Player Load was reset to zero at the start of each session, so no additional reset was required during sequence construction. Missing values~\cite{siddappa2026alert} occurred only in Medic~5 and Medic~8 and represented less than 2\% of the affected sessions; these values were handled using linear interpolation independently within each session and chronological partition.

HR forecasting was formulated as a supervised multivariate, multi-horizon time-series prediction problem. For each session, a window of recent wearable observations was used to predict the subsequent HR trajectory. Let \(W\) denote the input-window length and \(H\) the forecast horizon. For time index \(t\), the input sequence is defined as
\begin{equation}
\mathbf{X}_{t}
=
[
\mathbf{x}_{t},
\mathbf{x}_{t+1},
\ldots,
\mathbf{x}_{t+W-1}
],
\end{equation}
where \(\mathbf{x}_{t}\in\mathbb{R}^{4}\) contains HR, velocity, acceleration, and Player Load. The corresponding multi-step target sequence is defined as
\begin{equation}
\mathbf{y}_{t}
=
[
\mathrm{HR}_{t+W},
\mathrm{HR}_{t+W+1},
\ldots,
\mathrm{HR}_{t+W+H-1}
].
\end{equation}

Chronological partitioning was performed before sliding-window construction. For each held-out participant, observations were ordered by session and elapsed time and partitioned into the first 70\% for training, the subsequent 10\% for validation, and the final 20\% for testing. Sliding-window samples were then generated independently within each partition and session using a stride of 1~s. Consequently, no input window or forecast target crossed a train--validation, validation--test, or session boundary. For the 30~s task, \(W=90\)~s and \(H=30\)~s; for the 60~s task, \(W=120\)~s and \(H=60\)~s.

Feature normalization was performed using standard scaling fitted only on
the training sequences for each fold. The scaler was fitted by reshaping
the training input tensor across all time steps and features, and the same
scaler was then applied unchanged to the corresponding validation and test
sequences. This procedure prevented information from validation or test
data from influencing feature normalization. The scaler-fitting procedure
was matched to each training regime. For PFST, the scaler was fitted using
only the held-out participant's training split. For POT, the scaler was
fitted using the population training split. For PPFT, the scaler fitted
during population pretraining was reused during participant-specific
fine-tuning, validation, and testing. This maintained consistency between
the pretrained feature representation and the fine-tuning stage while
avoiding leakage from validation or test data.

\subsection{Forecasting Models}

We evaluated three neural forecasting architectures: a long short-term
memory (LSTM) network, an Attention-LSTM, and a convolutional neural
network--LSTM (CNN-LSTM). These models were selected to represent
sequence-learning backbones with different mechanisms for capturing
temporal structure in wearable HR and workload time series. Evaluating
multiple architectures allowed us to determine whether the effect of
training regime was consistent across model families.

\subsubsection{LSTM Model}

The LSTM model served as the recurrent baseline. Input features were first normalized using layer normalization and then passed through a two-layer LSTM encoder. The encoded sequence was subsequently normalized, and three complementary representations were extracted: the final hidden state, the mean hidden representation across the input window, and the last observed input vector. These representations were concatenated and passed through a multilayer fully connected forecast head to predict the future HR trajectory.

\subsubsection{Attention-LSTM Model}

The Attention-LSTM model extended the LSTM architecture by incorporating a MLP-based temporal attention mechanism. After layer normalization and two-layer LSTM encoding, attention weights~\cite{raju2026cirrguide, raju2026chexfuse} were computed over the most recent 30 time steps, allowing the model to emphasize the most relevant portion of the history window while reducing the influence of less informative earlier observations. The resulting attention context vector was concatenated with the final LSTM hidden state and the last observed input vector. This fused representation was then passed through a multilayer forecast head to generate the multi-horizon HR prediction.

\subsubsection{CNN-LSTM Model}

The CNN-LSTM combined residual temporal convolution with recurrent
sequence modeling. Following input layer normalization, residual
one-dimensional convolutional blocks with 64 channels, kernel size 5,
and dilation factors 1, 2, and 4 extracted local and progressively wider temporal patterns. The resulting representation was passed through a two-layer LSTM encoder and layer normalization. The final hidden state, an attention-derived context vector over the LSTM outputs, and the last observed input vector were concatenated and passed through a multilayer forecast head.

\subsubsection{Persistence Baseline}

A persistence baseline was included to account for the strong short-term continuity of HR. For each input window, the baseline repeated the last observed HR value across all future forecast steps:
\begin{equation}
\widehat{\mathrm{HR}}_{t+W+k-1}
=
\mathrm{HR}_{t+W-1},
\quad
k = 1,2,\ldots,H.
\end{equation}

This baseline is particularly important for short forecast horizons because HR changes gradually over time. Comparing neural models against persistence helps determine whether the models learn meaningful temporal-workload relationships beyond simple HR continuity.

\subsubsection{Decomposition-Based Fine-Tuning}
To test whether separating slow-varying trend from short-term residual improves forecasting, HR was decomposed independently within each chronologically ordered session using a causal exponentially weighted moving average (EWMA) with a 60~s span. The trend at time \(t\) used only HR observations up to \(t\), preventing future information leakage; the residual was
\begin{equation}
\mathrm{HR}^{\mathrm{residual}}_{t}
=
\mathrm{HR}_{t}
-
\mathrm{HR}^{\mathrm{trend}}_{t}.
\end{equation}

Separate trend ($M_{\mathrm{trend}}$) and residual ($M_{\mathrm{resid}}$) forecasting models used the corresponding direct model's architecture. Each predicted its respective component from that component plus velocity, acceleration, and Player Load. Both models were pretrained on non-held-out participants and fine-tuned on the held-out participant using the PPFT protocol. At inference, the component predictions were summed over forecast interval \(\mathcal{T}_{t}=\{t+W,\ldots,t+W+H-1\}\):
\begin{equation}
\widehat{\mathrm{HR}}_{\mathcal{T}_{t}}
=
\widehat{\mathrm{Trend}}_{\mathcal{T}_{t}}
+
\widehat{\mathrm{Residual}}_{\mathcal{T}_{t}}.
\end{equation}

\begin{figure*}[t]
\centering
\includegraphics[width=\textwidth]{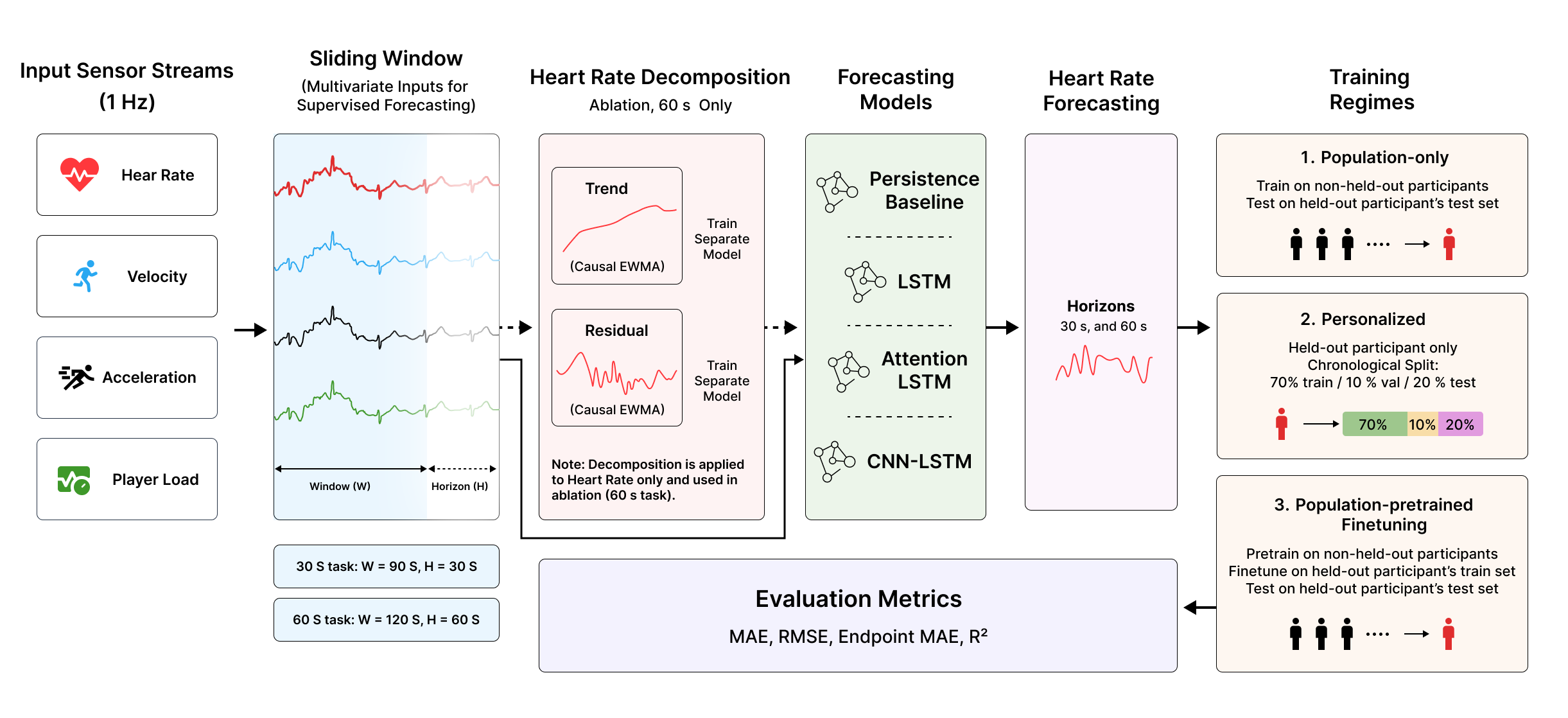}
\caption{Methodological overview of the wearable-based HR forecasting
benchmark for loaded ruck marching. Wearable inputs were converted into
supervised forecasting sequences and evaluated using the persistence
baseline, LSTM, Attention-LSTM, and CNN-LSTM models under POT, PFST,
and PPFT. The decomposition pathway was evaluated only as a secondary
60~s ablation.}
\label{fig:method_overview}
\end{figure*}

\subsection{Training Regimes}

The primary objective of this study was to evaluate how different
training strategies affect wearable-based HR forecasting under scarce
ruck-march data conditions. For each neural architecture, three training
regimes were compared: POT, PFST, and PPFT. Each regime was evaluated
within a leave-one-participant-out framework, in which one participant
served as the held-out participant and the remaining seven participants
formed the population source. The full protocol was repeated across all
participants, with each participant serving once as the held-out test
participant. Fig.~\ref{fig:method_overview} presents the overall
methodological overview.

\subsubsection{Population-Only Training (POT)}

In POT, one participant was held out for final testing, and all data from
that participant were excluded from population training and validation.
The remaining seven participants formed the population source. Within
each leave-one-participant-out fold, the seven non-held-out participants
were shuffled using a fixed split seed of 42, after which one participant
was assigned to validation and the remaining six were used for population
training. The resulting population train--validation split was held fixed
across all training seeds. After training, the population model was
evaluated on the final 20\% of the held-out participant's chronologically
ordered data. POT therefore measured cross-participant generalization,
representing the case in which a model trained on other participants is
applied to a new participant without participant-specific adaptation.

\subsubsection{Personalized-From-Scratch Training (PFST)}

In PFST, model training used only data from the held-out participant.
The participant's data were ordered chronologically by session and
elapsed time within each session. The first 70\% of the
participant-specific timeline was used for training, the next 10\% for
validation, and the final 20\% for testing. PFST represented direct
personalization under scarce-data conditions and tested whether a model
trained exclusively on limited participant-specific data could learn an
effective HR forecasting function without population-level pretraining.

\subsubsection{Population-Pretrained Fine-Tuning (PPFT)}

In PPFT, the seven non-held-out participants formed the population
source, with the same fold-specific population train--validation split
used for POT. The population checkpoint with the lowest validation loss was
then used to initialize the participant-specific model. The pretrained
model was subsequently fine-tuned using the first 70\% of the held-out
participant's chronologically ordered data, validated on the next 10\%,
and evaluated on the final 20\%. Fine-tuning was performed end-to-end,
with all model parameters updated and no layers frozen. A reduced
learning rate was used during fine-tuning to adapt the pretrained temporal
representation while limiting abrupt parameter updates.

The same held-out test segment was used for POT, PFST, and PPFT within
each leave-one-participant-out fold. This ensured that performance
differences reflected the training regime rather than differences in the
evaluation data.

\subsection{Model Training}

All neural forecasting models were trained using the same optimization
protocol to ensure a consistent comparison across architectures and
training regimes. Huber loss was used as the training objective, and model parameters were optimized using AdamW with a weight decay of
$10^{-4}$. A ReduceLROnPlateau scheduler reduced the learning rate by a
factor of 0.5 when validation loss stopped improving. Early stopping was
also based on validation loss, and the checkpoint with the lowest
validation loss was retained for final evaluation on the held-out test
data. For the LSTM, Attention-LSTM, and CNN-LSTM experiments, the recurrent hidden size was set to 160 with two recurrent layers. The LSTM and CNN-LSTM models used a dropout rate of 0.20, whereas the Attention-LSTM used a dropout rate of 0.25. POT and PFST used an initial learning rate of $7\times10^{-4}$, whereas PPFT used a smaller learning rate of $1\times10^{-4}$. A batch size of 64 was used for all experiments. POT and PFST were trained for a maximum of 50 epochs with an early-stopping patience of 10 epochs, while PPFT was performed for a maximum of 30 epochs with an early-stopping patience of 5 epochs.

Additionally, we evaluated PFST model-capacity sensitivity using three additional configurations with 32, 64, and 96 hidden units, 1--2
recurrent layers, 16--32 CNN channels, and 1--2 CNN blocks, alongside the primary 160-unit configuration. Within this sensitivity analysis, the best configuration for each participant was selected by validation loss. 

All main experiments were conducted using three training seeds: 42, 7, and 123. Metrics were first averaged across seeds within each participant. Paired differences between PPFT and each comparator (PFST, POT, and persistence) were then evaluated at the participant level using 10,000 percentile-bootstrap resamples to obtain 95\% confidence intervals and exact two-sided sign-flip tests over all $2^8=256$ possible sign assignments. Sign-flip $p$-values were adjusted for multiple comparisons using Holm's method within each comparator--metric family of six architecture--horizon tests.

\subsection{Evaluation Metrics}

Model performance was evaluated using mean absolute error (MAE), root mean squared error (RMSE), endpoint MAE, and coefficient of determination \(R^2\). MAE, RMSE, and \(R^2\) were computed over all predicted time steps by flattening the multi-horizon prediction matrix across forecast steps and supervised sequences. Endpoint MAE was computed only at the final forecast step, corresponding to 30~s ahead for the 30~s forecasting task and 60~s ahead for the 60~s forecasting task, and
therefore provides a complementary measure of accuracy at the maximum forecast lead. In addition, since \(R^2\) is sensitive to the variance of the HR signal, it was interpreted at the participant level alongside the test-set audit.

\section{Results}
\subsection{Data Scarcity and Test-Set Audit}
\label{sec:data_scarcity_results}

The scarce-data setting is defined here at the participant and session level: only eight participants and thirteen ruck-march sessions were available, and each participant contributed one or two sessions. Although the 1~s sliding-window procedure produced approximately $3.0\times10^4$ to $3.5\times10^4$ highly overlapping supervised sequences in the population folds, this nominal sequence count does not replace independent participant or session diversity. PFST and PPFT were restricted to the first 70\% of the held-out participant's limited chronological timeline for training, with the next 10\% used for validation and the final 20\% used for testing.
Population training sets contained approximately 530--617~min of
ruck-march data, whereas the 60~s held-out test segments ranged from 11.18 to 23.68~min and contained 492 to 1242 supervised sequences. Table~\ref{tab:test_audit_60s} summarizes the held-out test segments at the 60~s horizon. The lower Player Load means for Medics~2 and~5 reflect their held-out segments occurring within shorter, lower-velocity second sessions, where cumulative Player Load resets at session onset. HR variability differed substantially across participants: HR range varied from 18.0~bpm in Medic~4 to 80.9~bpm in Medic~3, and HR standard deviation ranged from 3.05 to 25.68~bpm. Because $R^2$ is sensitive to target variance, low-variance folds can yield low or negative $R^2$ despite modest absolute error. Accordingly, MAE, RMSE, and endpoint MAE were used for the primary aggregate comparison, whereas $R^2$ was treated as a participant-level diagnostic.

\begin{table*}[t]
\caption{Held-out test-set audit at the 60-s forecasting horizon.
Test segments comprise the final 20\% of each held-out participant's
chronologically ordered data. HR denotes heart rate; PL denotes player
load. Values are mean $\pm$ standard deviation.}
\label{tab:test_audit_60s}
\centering
\footnotesize

\begin{tabular}{lcccccc}
\hline
\textbf{Participant} &
\textbf{Duration (min)} &
\textbf{Sequences} &
\textbf{HR Mean (bpm)} &
\textbf{HR Range (bpm)} &
\textbf{PL Mean} &
\textbf{PL Range} \\
\hline

Medic 1 & 15.10 & 727
& $169.25 \pm 10.36$
& 36.0
& $525.24 \pm 31.71$
& 109.7 \\

Medic 2 & 20.17 & 1031
& $135.44 \pm 11.99$
& 51.0
& $26.17 \pm 9.33$
& 33.3 \\

Medic 3 & 11.18 & 492
& $143.75 \pm 25.68$
& 80.9
& $405.39 \pm 14.70$
& 58.3 \\

Medic 4 & 13.75 & 646
& $127.95 \pm 3.05$
& 18.0
& $308.79 \pm 13.52$
& 46.6 \\

Medic 5 & 19.98 & 1020
& $153.75 \pm 10.99$
& 46.0
& $30.85 \pm 11.27$
& 46.5 \\

Medic 6 & 23.40 & 1225
& $175.55 \pm 19.50$
& 64.4
& $290.04 \pm 50.29$
& 180.0 \\

Medic 7 & 23.68 & 1242
& $178.08 \pm 15.87$
& 58.0
& $308.63 \pm 55.68$
& 196.8 \\

Medic 8 & 22.38 & 1164
& $162.62 \pm 15.46$
& 61.0
& $299.75 \pm 42.62$
& 151.8 \\

\hline
\end{tabular}
\end{table*}

\subsection{Effect of Training Regime Under Scarce Individual Data}
\label{sec:training_regime_results}

Across all three neural architectures and both forecast horizons, PPFT
yielded lower mean MAE, RMSE, and endpoint MAE than both PFST and POT,
as reported in Tables~\ref{tab:aggregate_error_30s}
and~\ref{tab:aggregate_error_60s}. Relative to PFST, PPFT reduced mean
MAE by 45.7--54.7\% across architectures at 30~s and by 27.9--36.1\%
at 60~s. Relative to POT, the corresponding MAE reductions were
0.9--7.5\% at 30~s and 5.2--8.3\% at 60~s. The PFST model-capacity sensitivity analysis yielded only modest changes in forecasting MAE ($-0.275$ to $+0.114$~bpm) relative to the primary PFST configuration across horizons and neural backbones, indicating that architecture tuning did not substantially alter the performance gap between PFST and PPFT. Thus, the most consistent result of the benchmark was associated with training strategy rather than a single model backbone: population pretraining provided a stronger starting point for participant-specific adaptation than PFST under limited personal data.

\subsubsection{30-s Horizon}

At 30~s, persistence remained highly competitive, with
$3.555 \pm 1.274$~bpm MAE and $5.117 \pm 1.719$~bpm RMSE.
Persistence achieved the lowest overall MAE, while CNN-LSTM PPFT
achieved a closely comparable MAE of $3.567 \pm 1.533$~bpm and the
lowest RMSE of $4.669 \pm 1.916$~bpm. LSTM PPFT achieved the lowest
endpoint MAE at $5.093 \pm 2.365$~bpm. The 0.012~bpm difference
between persistence and the lowest PPFT MAE indicates that
trajectory-averaged error at 30~s remained strongly influenced by
short-term HR continuity. The lower RMSE and endpoint MAE achieved by
PPFT suggest greater benefit from learned temporal--workload structure
for larger deviations and toward the end of the forecast interval.

\begin{table*}[t]
\caption{Aggregate forecasting error at the 30-s horizon across three training
seeds (42, 7, and 123), reported as mean $\pm$ standard deviation across
held-out participants after averaging across seeds within each participant.}
\label{tab:aggregate_error_30s}
\centering
\footnotesize

\begin{tabular}{llccc}
\hline
\textbf{Model} &
\textbf{Training Regime} &
\textbf{MAE (bpm)} &
\textbf{RMSE (bpm)} &
\textbf{Endpoint MAE (bpm)} \\
\hline

Baseline & Persistence
& \textbf{$3.555 \pm 1.274$}
& $5.117 \pm 1.719$
& $6.020 \pm 2.108$ \\
\hline

LSTM & POT
& $3.663 \pm 1.451$
& $4.818 \pm 1.872$
& $5.248 \pm 2.327$ \\

LSTM & PFST
& $6.688 \pm 3.069$
& $8.384 \pm 3.595$
& $7.053 \pm 3.277$ \\

LSTM & PPFT
& $3.629 \pm 1.573$
& $4.750 \pm 1.934$
& \textbf{$5.093 \pm 2.365$} \\
\hline

Attention-LSTM & POT
& $3.760 \pm 1.488$
& $4.919 \pm 1.904$
& $5.301 \pm 2.315$ \\

Attention-LSTM & PFST
& $6.986 \pm 3.686$
& $8.564 \pm 4.096$
& $7.226 \pm 3.618$ \\

Attention-LSTM & PPFT
& $3.593 \pm 1.573$
& $4.738 \pm 1.935$
& $5.094 \pm 2.342$ \\
\hline

CNN-LSTM & POT
& $3.855 \pm 1.392$
& $4.999 \pm 1.808$
& $5.368 \pm 2.306$ \\

CNN-LSTM & PFST
& $7.868 \pm 3.926$
& $9.507 \pm 4.353$
& $7.987 \pm 4.007$ \\

CNN-LSTM & PPFT
& $3.567 \pm 1.533$
& \textbf{$4.669 \pm 1.916$}
& $5.107 \pm 2.333$ \\
\hline
\end{tabular}

\end{table*}

\subsubsection{60-s Horizon}

Forecasting became more difficult at 60~s: persistence MAE increased to
$5.619 \pm 1.797$~bpm and endpoint MAE to
$8.735 \pm 2.802$~bpm. At this horizon, Attention-LSTM PPFT achieved
the lowest aggregate MAE and RMSE, at $5.573 \pm 1.939$ and
$7.134 \pm 2.260$~bpm, respectively, while CNN-LSTM PPFT achieved
the lowest endpoint MAE at $7.553 \pm 2.357$~bpm. Across all three
architectures, PPFT yielded lower MAE, RMSE, and endpoint MAE than the
corresponding POT and PFST regimes. The larger separation between PFST
and PPFT at 60~s further supports the central scarce-data result that
participant-specific adaptation was more effective when initialized from a
population-trained model. 

\begin{table*}[t]
\caption{Aggregate forecasting error at the 60-s horizon across three training
seeds (42, 7, and 123), reported as mean $\pm$ standard deviation across
held-out participants after averaging across seeds within each participant.}
\label{tab:aggregate_error_60s}
\centering
\footnotesize

\begin{tabular}{llccc}
\hline
\textbf{Model} &
\textbf{Training Regime} &
\textbf{MAE (bpm)} &
\textbf{RMSE (bpm)} &
\textbf{Endpoint MAE (bpm)} \\
\hline

Baseline & Persistence
& $5.619 \pm 1.797$
& $7.833 \pm 2.531$
& $8.735 \pm 2.802$ \\
\hline

LSTM & POT
& $6.046 \pm 1.858$
& $7.674 \pm 2.214$
& $8.089 \pm 2.644$ \\

LSTM & PFST
& $7.951 \pm 3.100$
& $9.985 \pm 3.726$
& $8.908 \pm 2.732$ \\

LSTM & PPFT
& $5.733 \pm 1.958$
& $7.302 \pm 2.330$
& $7.673 \pm 2.445$ \\
\hline

Attention-LSTM & POT
& $5.914 \pm 1.629$
& $7.551 \pm 1.940$
& $8.102 \pm 2.279$ \\

Attention-LSTM & PFST
& $8.310 \pm 3.811$
& $10.421 \pm 4.250$
& $9.009 \pm 3.251$ \\

Attention-LSTM & PPFT
& \textbf{$5.573 \pm 1.939$}
& \textbf{$7.134 \pm 2.260$}
& $7.620 \pm 2.314$ \\
\hline

CNN-LSTM & POT
& $6.194 \pm 2.013$
& $7.816 \pm 2.381$
& $8.233 \pm 2.655$ \\

CNN-LSTM & PFST
& $8.887 \pm 3.366$
& $11.064 \pm 4.077$
& $9.916 \pm 3.675$ \\

CNN-LSTM & PPFT
& $5.683 \pm 1.948$
& $7.313 \pm 2.279$
& \textbf{$7.553 \pm 2.357$} \\
\hline
\end{tabular}

\end{table*}

\subsubsection{Participant-Level Paired Analysis}
Participant-level paired analysis after averaging across the three
training seeds further distinguished the training regimes. Relative to
PFST, PPFT reduced trajectory MAE across all architectures and horizons.
At 30~s, the reductions were 3.059~bpm (95\% CI: 2.038--4.412) for
LSTM, 3.393~bpm (1.995--5.027) for Attention-LSTM, and 4.301~bpm
(2.624--6.192) for CNN-LSTM. At 60~s, the corresponding reductions
were 2.219~bpm (0.921--3.541), 2.737~bpm (1.352--4.292), and
3.204~bpm (1.910--4.603), respectively. All six trajectory-MAE
comparisons remained significant after Holm adjustment
($p_{\mathrm{adj}}=0.0469$).
Endpoint-MAE reductions were 1.961~bpm (0.973--3.334),
2.132~bpm (1.201--3.234), and 2.879~bpm (1.596--4.304) at
30~s, and 1.235~bpm (0.349--2.155), 1.389~bpm
(0.339--2.701), and 2.364~bpm (1.037--3.998) at 60~s.
Four of the six endpoint-MAE comparisons remained significant after
Holm adjustment; the 60-s LSTM and Attention-LSTM comparisons had
$p_{\mathrm{adj}}=0.0781$.

Relative to POT, PPFT produced smaller improvements. Trajectory-MAE
reductions ranged from 0.03 to 0.51~bpm, and none of the six
corresponding comparisons remained significant after Holm adjustment.
Endpoint-MAE reductions ranged from 0.16 to 0.68~bpm, with two of six
comparisons remaining significant after Holm adjustment, corresponding
to CNN-LSTM at 30~s and 60~s ($p_{\mathrm{adj}}=0.0469$). Thus, the strongest effect of PPFT was observed relative to PFST, while the additional improvement over POT was more modest.

Relative to persistence, trajectory-MAE differences were small,
consistent with the strong short-horizon baseline. In contrast, PPFT
reduced endpoint MAE by 0.91--1.18~bpm across architectures and
horizons. Endpoint MAE was lower with PPFT than with persistence in
7 of 8 participants in each architecture--horizon comparison. All six endpoint-MAE comparisons were significant at the unadjusted level ($p<0.05$); however, none remained significant after Holm adjustment ($p_{\mathrm{adj}}=0.1875$). 

\subsection{Participant-Level $R^2$ and Variance Sensitivity}
\label{sec:participant_r2_results}

Participant-level $R^2$ was examined as a secondary diagnostic of
participant-specific behavior. At 30~s, $R^2$ was generally high for
participants with substantial held-out HR variation, and PPFT typically
recovered or improved variance explanation relative to PFST. At 60~s,
$R^2$ became more heterogeneous. PPFT generally improved $R^2$ relative
to PFST, particularly for participants in which PFST produced low or
negative values. For example, for Medic~5, LSTM $R^2$ increased from
$-0.940$ with PFST to $0.360$ with PPFT, while CNN-LSTM increased from
$-1.237$ to $0.310$. POT also transferred well to several participants,
indicating shared cross-participant HR--workload structure, while PPFT
provided additional participant-specific adaptation. Participant-level
60-s $R^2$ results, averaged across the three training seeds, are reported in Table~\ref{tab:participant_r2_60s}.

\begin{table*}[t]
\caption{Participant-level overall $R^2$ at the 60-s forecasting horizon,
after averaging across three training seeds (42, 7, and 123) within each
participant.}
\label{tab:participant_r2_60s}
\centering
\footnotesize
\setlength{\tabcolsep}{3pt}
\renewcommand{\arraystretch}{1.10}

\begin{minipage}{0.88\textwidth}
\centering

\begin{tabular*}{\linewidth}{@{\extracolsep{\fill}}lcccccccccc}
\hline
\textbf{Participant} &
\textbf{Persistence} &
\multicolumn{3}{c}{\textbf{LSTM}} &
\multicolumn{3}{c}{\textbf{Attention-LSTM}} &
\multicolumn{3}{c}{\textbf{CNN-LSTM}} \\
\cline{3-5} \cline{6-8} \cline{9-11}

& &
\textbf{POT} & \textbf{PFST} & \textbf{PPFT} &
\textbf{POT} & \textbf{PFST} & \textbf{PPFT} &
\textbf{POT} & \textbf{PFST} & \textbf{PPFT} \\
\hline

Medic 1
& 0.738
& 0.682 & 0.536 & 0.791
& 0.727 & 0.486 & 0.775
& 0.636 & 0.479 & 0.742 \\

Medic 2
& 0.127
& 0.164 & -0.493 & 0.234
& 0.176 & -0.648 & 0.207
& 0.048 & -0.467 & 0.213 \\

Medic 3
& 0.651
& 0.799 & 0.486 & 0.817
& 0.802 & 0.475 & 0.817
& 0.795 & 0.411 & 0.792 \\

Medic 4$^{\mathrm{a}}$
& -0.297
& -0.488 & -1.786 & -0.249
& -1.453 & -0.974 & -0.330
& -0.615 & -1.116 & -0.237 \\

Medic 5
& 0.583
& 0.416 & -0.940 & 0.360
& 0.386 & -1.466 & 0.343
& 0.323 & -1.237 & 0.310 \\

Medic 6
& 0.807
& 0.866 & 0.862 & 0.852
& 0.876 & 0.866 & 0.881
& 0.880 & 0.801 & 0.873 \\

Medic 7
& 0.782
& 0.741 & 0.661 & 0.811
& 0.770 & 0.659 & 0.832
& 0.776 & 0.286 & 0.827 \\

Medic 8
& 0.614
& 0.601 & 0.642 & 0.632
& 0.677 & 0.487 & 0.690
& 0.579 & 0.522 & 0.669 \\
\hline
\end{tabular*}

\par
\raggedright\footnotesize
$^{\mathrm{a}}$ Medic~4 had a low-variance held-out HR segment;
therefore, $R^2$ is particularly sensitive to small absolute errors.\par

\end{minipage}
\end{table*}

Medic~4 remained the clearest variance-sensitive exception. Its held-out HR range was only 18.0~bpm with a standard deviation of 3.05~bpm; therefore, relatively small trajectory deviations were amplified in $R^2$. This participant illustrates why $R^2$ should be interpreted alongside absolute-error metrics and why MAE and RMSE were used as the primary aggregate measures in this small heterogeneous cohort.

\subsection{Decomposition-Based Fine-Tuning Ablation}
\label{sec:decomposition_results}
As a secondary analysis, a decomposition-based PPFT ablation was evaluated at 60~s using seed~42 to test whether explicit separation
of HR into causal trend and residual components improved adaptation.
As shown in Table~\ref{tab:decomp_error_60s}, direct PPFT produced lower
mean MAE, RMSE, and endpoint MAE than decomposition-based PPFT for LSTM
and Attention-LSTM. For CNN-LSTM, decomposition slightly reduced MAE
from $5.463 \pm 1.745$ to $5.416 \pm 1.984$~bpm, while RMSE and
endpoint MAE were slightly higher. Thus, decomposition provided no
consistent aggregate error advantage over direct PPFT.

\begin{table*}[t]
\caption{Aggregate comparison of direct and decomposition-based PPFT at the
60-s forecasting horizon for seed~42. Values are mean $\pm$ standard
deviation across held-out participants.}
\label{tab:decomp_error_60s}
\centering
\footnotesize

\begin{tabular}{llccc}
\hline
\textbf{Model} &
\textbf{Fine-Tuning Approach} &
\textbf{MAE (bpm)} &
\textbf{RMSE (bpm)} &
\textbf{Endpoint MAE (bpm)} \\
\hline

LSTM & Direct PPFT
& $5.532 \pm 1.908$
& $7.135 \pm 2.278$
& $7.551 \pm 2.515$ \\

LSTM & Decomposition PPFT
& $5.620 \pm 2.197$
& $7.272 \pm 2.633$
& $8.121 \pm 3.191$ \\
\hline

Attention-LSTM & Direct PPFT
& $5.407 \pm 1.991$
& $6.964 \pm 2.271$
& $7.379 \pm 2.243$ \\

Attention-LSTM & Decomposition PPFT
& $5.633 \pm 1.944$
& $7.134 \pm 2.254$
& $7.647 \pm 2.605$ \\
\hline

CNN-LSTM & Direct PPFT
& $5.463 \pm 1.745$
& $7.032 \pm 2.051$
& $7.439 \pm 2.241$ \\

CNN-LSTM & Decomposition PPFT
& $5.416 \pm 1.984$
& $7.060 \pm 2.399$
& $7.707 \pm 2.600$ \\
\hline
\end{tabular}

\end{table*}

Participant-level $R^2$ showed a similarly mixed pattern. Decomposition
improved $R^2$ in seven of eight LSTM folds, but in only three of eight
Attention-LSTM folds and three of eight CNN-LSTM folds. The effect was
therefore participant- and architecture-dependent rather than systematic.

\section{Discussion}

The central finding of this study is that scarce-data ruck-march HR
forecasting is better approached through PPFT than through PFST. Across
architectures and forecast horizons, population pretraining provided a
stronger initialization for participant-specific adaptation than learning
the forecasting function from limited participant-specific data alone.
This pattern suggests that HR responses during loaded ruck marching
contain transferable cross-participant structure that can be learned
through POT and subsequently refined through PPFT.

The architecture results further support this interpretation. No single neural backbone was uniformly superior across metrics and forecast horizons, whereas the direction of the training-regime results was consistent across LSTM, Attention-LSTM, and CNN-LSTM. Thus, within this benchmark, the choice of training regime appeared to be more consistent than the choice of neural backbone. 
The smaller difference between PPFT and POT provides additional insight
into the role of participant-specific adaptation. POT remained
competitive when error was averaged across the full forecast trajectory,
indicating that substantial transferable HR--workload structure was
already captured before participant-specific adaptation. However, PPFT
consistently achieved lower endpoint MAE than POT across architectures
and horizons. This pattern suggests that participant-specific adaptation
provided greater additional benefit toward the maximum forecast lead,
where individual differences in physiological response may become more
important to the forecast.

The persistence baseline reinforces this forecast-lead-dependent
interpretation. Persistence remained highly competitive for
trajectory-averaged error, particularly at the 30~s horizon, reflecting
the strong short-term continuity of HR. In contrast, the relative benefit
of PPFT became more apparent toward the maximum forecast lead and at the
longer forecasting horizon, where repeating the most recently observed HR
was less effective. These findings suggest that learned HR--workload
dynamics may provide greater value for predictions farther into the
forecast horizon than for the earliest forecast steps.
Participant-level $R^2$ remained heterogeneous and sensitive to the
variance of the held-out HR segment. PPFT generally improved $R^2$
relative to PFST, particularly when PFST produced low or negative values,
while the low-variance Medic~4 test segment illustrated how relatively
small absolute errors can produce unfavorable $R^2$ values. This
heterogeneity supports the use of MAE and RMSE as the primary aggregate
metrics while retaining $R^2$ as a participant-level diagnostic.

Finally, the decomposition-based PPFT ablation provided no consistent
aggregate advantage over direct PPFT. Explicitly separating HR into trend
and residual components therefore did not provide a clear additional
benefit under the present experimental setting.

\section{Limitations and Future Work}
The proposed generalize-then-personalize strategy demonstrates promising performance for wearable HR forecasting under scarce ruck-march data conditions. However, the present benchmark was conducted using eight participants and thirteen ruck-march sessions, with each participant contributing one or two sessions. This focused dataset reflects the practical constraints associated with collecting ruck-march data and therefore represents the scarce-data setting motivating this study. Nevertheless, evaluation on larger and more diverse cohorts is needed to assess the broader generalizability of the findings. In addition, the amount of participant-specific data was constrained by the available ruck-march data rather than systematically varied across different adaptation-data levels. The present results should therefore be interpreted as descriptive evidence within the practically constrained, data-scarce ruck-march setting examined in this study. Future work should extend this benchmark to larger and more diverse ruck-march datasets with additional participants and repeated sessions per participant.

\section{Conclusion}

This study presented, to the best of our knowledge, the first controlled
benchmark of population-only training (POT), personalized-from-scratch
training (PFST), and population-pretrained fine-tuning (PPFT) for
wearable HR forecasting during loaded ruck marching under scarce-data
conditions. Across neural backbones and forecast horizons, PPFT yielded
lower mean MAE, RMSE, and endpoint MAE than the corresponding PFST and
POT regimes. The largest improvements were observed relative to PFST,
supporting population pretraining as a stronger initialization for
participant-specific adaptation than learning the forecasting function
from limited participant-specific data alone. POT remained competitive
for trajectory-averaged forecasting, indicating substantial transferable
cross-participant HR--workload structure, while PPFT provided a clearer
additional benefit toward the maximum forecast lead. Persistence was also
highly competitive for trajectory-averaged error, particularly at 30~s,
but PPFT consistently achieved lower endpoint MAE across architectures
and horizons. The findings
support a generalize-then-personalize strategy in which transferable
cross-participant HR--workload structure is first learned through
population pretraining and subsequently refined through
participant-specific adaptation.

\section*{Acknowledgment}
This study was funded by the Department of Defense via the Harnessing Emerging Research Opportunities to Empower Soldiers (HEROES) program. OPSEC PR2026-3942

\bibliographystyle{IEEEtran}
\bibliography{references}
\end{document}